\documentclass{article}
\usepackage[utf8]{inputenc}
\usepackage{iclr2027_conference,times}
\usepackage{booktabs}

\usepackage{amsmath,amsfonts,bm}

\def\eqref#1{equation~\ref{#1}}

\def\1{\bm{1}}

\DeclareMathAlphabet{\mathsfit}{\encodingdefault}{\sfdefault}{m}{sl}
\SetMathAlphabet{\mathsfit}{bold}{\encodingdefault}{\sfdefault}{bx}{n}

\usepackage{hyperref}
\hypersetup{hidelinks}
\usepackage{url}
\usepackage{amsmath}
\usepackage{natbib}
\usepackage{graphicx}
\usepackage{booktabs}
\usepackage{adjustbox}
\usepackage{float}
\usepackage{capt-of}
\usepackage{booktabs}
\usepackage{graphicx}
\usepackage{pifont}
\usepackage{wrapfig}
\usepackage{amssymb,tabularx,fancyvrb}
\newsavebox{\realworldbox}
\newlength{\realworldheight}
\newlength{\realworldcolwidth}

\title{PointPilot: Unified Robotic Manipulation via Object-Centric 3D Point Flow Prediction}

\newcommand{\PPauthor}[2]{\mbox{\textbf{#1}\textsuperscript{#2}}}
\newcommand{\PPaffil}[2]{\mbox{\textsuperscript{#1}\,#2}}

\newcommand{\PPauthorblock}{%
  \begin{minipage}{0.96\textwidth}
    \centering\normalfont
    \vspace*{1em}
    {
      \PPauthor{Yihan Fang}{1,2*}\quad
      \PPauthor{Yuxin Tian}{1,2*}\quad
      \PPauthor{Ruixiang Wang}{1,2}\quad
      \PPauthor{Zhanqi Zhang}{2}\quad
      \PPauthor{Xing Hu}{2}\quad
      \PPauthor{Zhen Yang}{1,2}\quad
      \PPauthor{Huayi Zhou}{3}\quad
      \PPauthor{Yongyi Su}{4}\quad
      \PPauthor{Wei Gao}{2}\quad
      \PPauthor{Kui Jia}{1,2}\par}
    \vspace{0.4em}
    {\small
      \PPaffil{1}{The Chinese University of Hong Kong, Shenzhen}\quad
      \PPaffil{2}{DexForce Technology Co., Ltd.}\par
      \PPaffil{3}{Shenzhen University}\quad
      \PPaffil{4}{Foshan University}\par}
    \vspace{0.2em}
    {\footnotesize \textsuperscript{*}\,Equal contribution.\par}
  \end{minipage}%
}

\author{\PPauthorblock}

\iclrfinalcopy 
\begin{document}

\maketitle
\fancyhead[L]{}
\begin{abstract}
Generalization remains a central challenge in robotics, calling for policies that address domain variations from multiple dimensions such as diverse embodiments, camera poses, and/or scene layouts.
In order to approach this problem, we introduce \textbf{PointPilot}, a task-focused and 3D aligned manipulation framework built on a 3D space point-flow prediction.
Rather than modeling the global scene, PointPilot only represents task-focused objects (robot grippers treated as objects), 
avoiding the prediction of potentially redundant scene parts unrelated to the manipulation tasks.
Moreover, 
PointPilot aligns 3D spatial representation in one unified coordinate and supports
closed-loop and fully automatic action execution across different manipulation modes from object point flow without a separately trained action expert, while existing works typically need an extra action head or are (semi-)automatic and single-arm mode specific.
Given sensor observations and language instructions,
our model jointly learns subtask planning, task-relevant
object grounding, and 3D point flow prediction in an end-to-end way.
By processing heterogeneous data in a shared 3D point flow space, PointPilot is able to absorb demonstrations from different embodiments.
Extensive simulative and real world experiments validate our method.
\emph{Notably}, we demonstrate the zero-shot embodiment generalization in the PointPilot --- one trained policy can be deployed onto previously unseen embodiments and successfully complete manipulation tasks, without any additional fine-tuning.
\end{abstract}
\vspace{-5pt}

\par\vspace{-5pt}

\section{Introduction}
    A central promise of generalist robot learning is that manipulation policies
should improve as they absorb increasingly diverse data. However, realizing this
promise requires more than scaling model parameters and demonstration
hours. Robot data is heterogeneous not only in visual appearance, task
semantics, and environment configuration, but also in embodiments. Native actions from heterogeneous data lack a shared representation form, preventing action prediction from directly serving  as a unified supervision space. The choice of prediction space
therefore determines whether heterogeneous demonstrations provide compatible
supervision or remain isolated by their control interfaces.

Vision--language--action(VLA) models have made substantial progress toward pretraining
a single policy across tasks and robot platforms
\citep{openx2024rtx,ghosh2024octo,black2025pi0} and finetuning with a specific task. Their apparent unification
primarily occurs at the model level, for the reason that learned targets remain robot actions,
whose dimensionality, coordinate convention, and control semantics depend on
the embodiment and manipulation setup. Existing systems accommodate these
differences through data pre-processing, embodiment
conditioning, or specialized action heads \citep{openx2024rtx,wang2024hpt,
doshi2025crossformer,zheng2025uniact,zheng2026xvla}.

\begin{figure}
    \centering
    \includegraphics[width=0.95\linewidth]{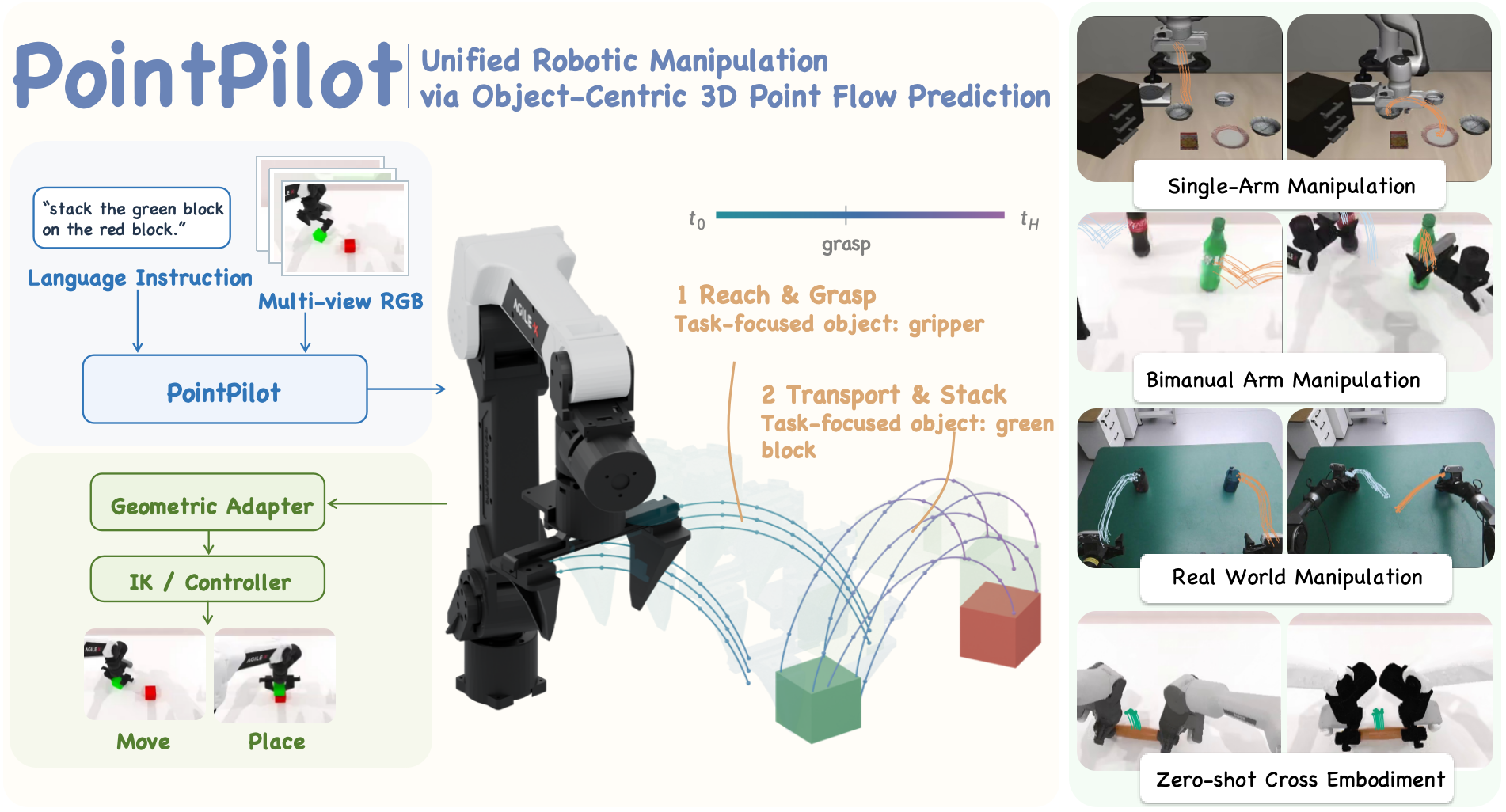}
    \par\vspace{-3mm}
    \caption{\textbf{Overview of PointPilot.} \textbf{Left:} Given a language instruction and multi-view RGB observations, PointPilot predicts task-focused objects' 3D point flow in each subtask. A learning-free adapter converts the predicted flow into robot end-effector pose commands for execution under the coupling condition. \textbf{Right:} PointPilot unifies in different manipulation modes and embodiments, thus supporting single-arm and bimanual gripper to execute any manipulation tasks by object-centric prediction. We further demonstrate real-world manipulation and promising zero-shot cross-embodiment skills transfer.}
    \label{fig:Overview}
    \vspace{-10pt}
\end{figure}

Task-relevant object motion offers a natural source of shared supervision because it captures task motion without committing to a robot's joints or visual appearance. This insight has motivated a growing family of representations, including articulation flow, general 3D object flow, object-part scene flow,
dense 3D motion fields, language-conditioned point trajectories, and point-cloud
world models \citep{eisner2022flowbot3d,yuan2025generalflow,xu2025im2flow2act,tang2025objectpart,yin2025motionfield,zhang2026molmomotion,huang2025particleformer,lin2026roboflow4d}.

Despite these advances in object-centric prediction for robotic manipulation, a unified and scalable framework remains challenging. To the best of our knowledge, existing approaches generally fall into
three patterns: \textbf{a)} Without any 3D alignment, they only use object motion as an intermediate latent feature for an additional action decoder, thus hardly processing the cross-embodiment generalization\citep{xu2025im2flow2act,yang2025ppi,lin2026roboflow4d}; \textbf{b)} depend on manual manipulation or an extra controller to grasp objects at the beginning, causing a modular framework and only support an open-looped or semi-automatic execution \citep{tang2025objectpart,yin2025motionfield,yuan2025generalflow}; \textbf{c)} or specialize object-centric representations to one particular
manipulation mode \citep{eisner2022flowbot3d, lin2026roboflow4d, tang2025objectpart, yang2025ppi, yuan2025generalflow}. \textbf{In particular}, manipulation modes such as single-arm or bimanual arm manipulation may not be a problem in VLA paradigm, but it is a problem in the
object-centric prediction. Unlike VLA, different manipulation modes require different numbers of object flows and different object-to-gripper motion assignments. To simplify this, most object-centric methods specify their flow to one object and only execute single-arm tasks.

On the other hand, an ideal
unified and scalable object-centric prediction framework should be \textbf{manipulation-mode agnostic}, allowing a single model to handle
both single-arm and dual-arm manipulation by predicting motion for a variable number of objects;
\textbf{embodiment-agnostic}, allowing heterogeneous robot demonstrations to be composed
despite incompatible native action spaces; and \textbf{directly executable}, allowing the object flow
prediction to control the robot in a closed loop without an additional embodiment-dependent
action decoder. Existing object-centric
approaches typically satisfy only a subset of these requirements, leaving object
motion either mode-specific, embodiment-dependent, or disconnected from fully automatic execution.

To alleviate this gap, \textbf{1)} we introduce a novel manipulation formulation that predicts task-focused future states and robot actions, in a unified 3D point-flow representation of relevant objects. 
Notable, we treat robot gripper tools as special objects and use the proposed 3D point-flow as an action representation, which accommodates different manipulation modes (e.g., single-arm or bimanual) and robot embodiments (e.g., parallel-jaw or dexterous hand) without platform-specific action experts.
Moreover, \textbf{2)} we align the 3D coordinates for both multi-view input observations and predicted 3D point flows in a shared reference frame. This alignment establishes cross-view spatial correspondences for visual and geometric fusion while providing a consistent prediction space across robot embodiments. 
Additionally, \textbf{3)} we integrate task understanding, semantic grounding, 3D point flow prediction and action generation into one model. This enables end-to-end training and fully autonomous closed-loop rollout, compared to various related works \citep{duan2024manipulate, huang2024copa, huang2024rekep, pan2025omnimanip, tang2025objectpart, 
yuan2025generalflow} that rely on external modules (e.g., sub-task orchestration or grasp planning) during the real world deployment. 
With these principles, \textbf{4)} we develop a comprehensive data processing pipeline, design the PointPilot model architecture, perform jointly training on heterogeneous robot data sources, and conduct various simulation and real world experiments.
PointPilot demonstrates promising generalization and scalability, such as zero-shot cross embodiment transfers. 
\par\vspace{-5pt}
\section{Related Work}
\label{sec:related_work}

\newcommand{\cmark}{\ding{51}}
\newcommand{\xmark}{\ding{55}}

\begin{wraptable}{r}{0.50\textwidth}
\centering
\vspace{-20pt}
\caption{\textbf{Comparison with prior object-centric methods.} Feat.1 means zero-shot cross-embodiment transfer. Feat.2 means no separate grasping module. Feat.3 means no learned motion-to-action action head. Feat.4 means single-arm manipulation tasks. Feat.5 means bimanual manipulation tasks.}
\label{tab:flow_method_comparison}
\setlength{\tabcolsep}{5pt}
\renewcommand{\arraystretch}{1.1}
\resizebox{\linewidth}{!}{%
\begin{tabular}{@{}lccccc@{}}
\toprule
Method & Feat.1 & Feat.2 & Feat.3 & Feat.4 & Feat.5 \\
\midrule
Im2Flow2Act \citep{xu2025im2flow2act}
& \xmark & \cmark & \xmark & \cmark & \xmark \\

PPI \citep{yang2025ppi}
& \xmark & \cmark & \xmark & \xmark & \cmark \\

RoboFlow4D \citep{lin2026roboflow4d}
& \xmark & \cmark & \xmark & \cmark & \xmark \\

General Flow \citep{yuan2025generalflow}
& \xmark & \xmark & \cmark & \cmark & \xmark \\

Object-Part Scene Flow \citep{tang2025objectpart}
& \cmark & \xmark & \cmark & \cmark & \xmark \\

3D Motion Field \citep{yin2025motionfield}
& \xmark & \xmark & \cmark & \cmark & \xmark \\

FlowBot3D \citep{eisner2022flowbot3d}
& \xmark & \cmark & \cmark & \cmark & \xmark \\

ImaginationPolicy \citep{lu2025imaginationpolicy}
& \cmark & \xmark & \cmark & \cmark & \xmark \\

\midrule
\textbf{PointPilot (Ours)}
& \cmark & \cmark & \cmark & \cmark & \cmark \\
\bottomrule
\end{tabular}%
}
\vspace{-15pt}
\end{wraptable}

\subsection{Object-Centric Motion Representations for Manipulation}
\vspace{-5pt}
Prior works have shown that
object-centric 3D flow provides a better generalization across unseen
objects, scenes, and robot embodiments
\citep{eisner2022flowbot3d,tang2025objectpart,
yuan2025generalflow,yin2025motionfield}.

However, existing object-centric approaches broadly follow several patterns.
One line treats predicted flow or point trajectories as intermediate
motion guidance for downstream action expert training
\citep{xu2025im2flow2act,yang2025ppi,
lin2026roboflow4d,zhang2026molmomotion}.
One line directly converts predicted objects' flow into robot
trajectories, but does not provide a complete autonomous and closed-looped manipulation. They either manually manipulate the gripper or use an extra controller to grasp the object.
\citep{bharadhwaj2024track2act,yuan2025generalflow,zhi2025flowaction}.
Another line is also often confined to a single manipulation mode,
such as the single-arm manipulation
\citep{eisner2022flowbot3d,yin2025motionfield,
yuan2025generalflow,zhi2025flowaction,lu2025imaginationpolicy,lin2026roboflow4d}. Most works only execute single-arm tasks for the reason that only one object needs to be predicted. The detailed comparison is shown in Table \ref{tab:flow_method_comparison}. Although RoboFlow4D is not a typical object-centric method, we include it because it represents gripper motion as point flow.
\subsection{Cross-Embodiment Policy Transfer}
\label{sec:related_cross_embodiment}
\vspace{-5pt}
Cross-embodiment policies use embodiment conditioning or shared latent
actions to learn from heterogeneous data~\citep{zheng2026xvla,bu2025univla}.
Object-centric 3D flow also supports human-to-robot
transfer~\citep{yuan2025generalflow} and cross-platform
deployment~\citep{zhi2025flowaction}.
PointPilot separates task-focused 3D motion prediction from robot-specific
execution: a non-learned geometric adapter maps predicted flows to
end-effector targets under local rigid coupling.
We demonstrate deployment on unseen embodiments without target-robot
demonstrations or fine-tuning, using each robot's kinematics and controller.

\par\vspace{-5pt}
\section{PointPilot: Unified Object-centric 3D Point Flow Prediction}
\vspace{-5pt}
Building on the definition of \emph{task-focused objects}, we introduce
\emph{PointPilot}. In this section, we will present the model architecture, covering subtask planning, object grounding, 3D geometric alignment, cross-view fusion , and point-flow predicting, and then describe how the predicted point flow is mapped to executable end-effector trajectories through a learning-free adapter.

\begin{figure}
    \centering
    \includegraphics[width=\linewidth]{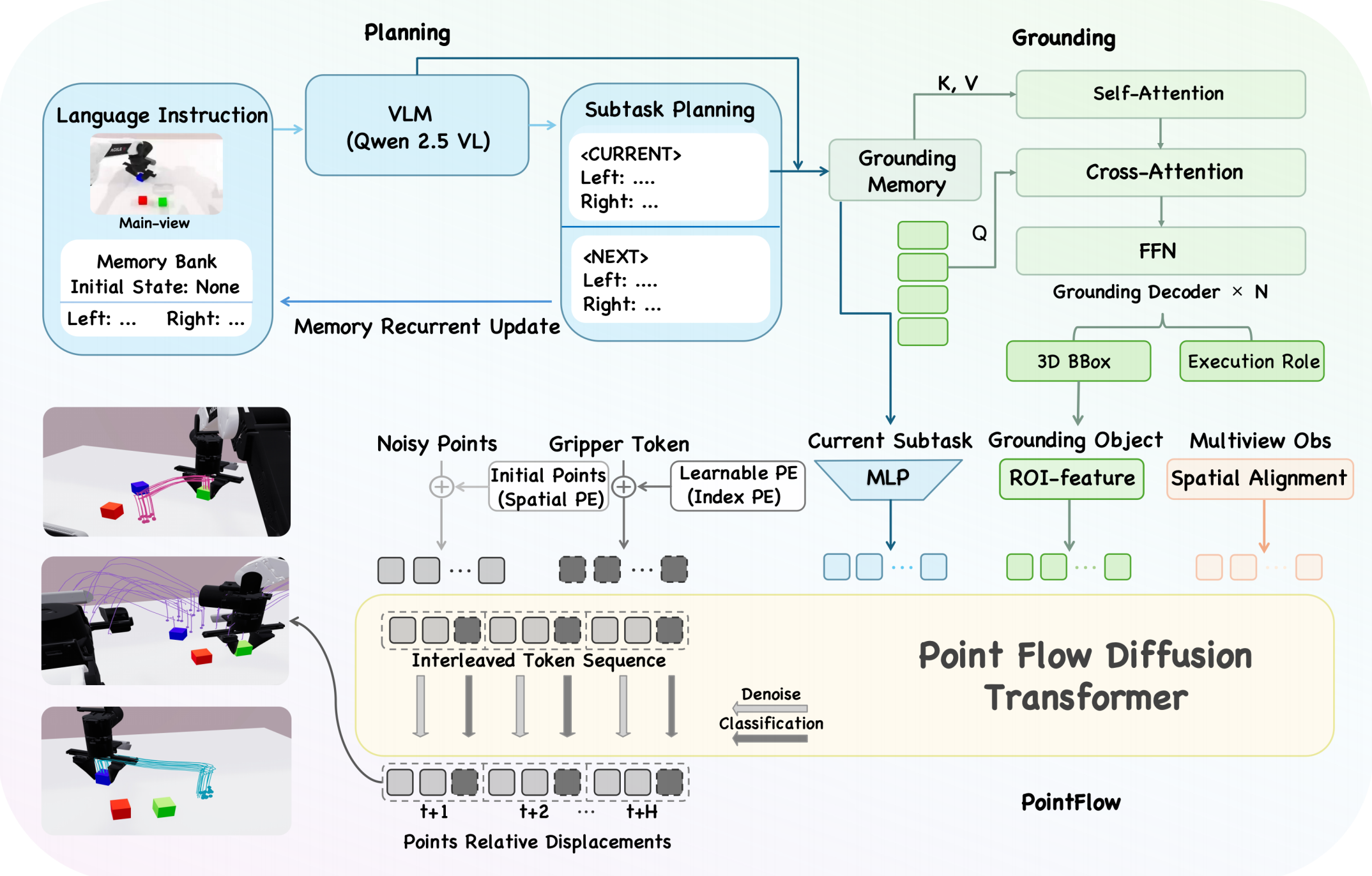}
    \par\vspace{-3mm}
    \caption{
\textbf{Method of PointPilot}.
\textbf{(a) Planning}: A memory-conditioned VLM predicts current and next subtasks.
\textbf{(b) Grounding}: Task-conditioned 3D grounding localizes task-focused objects and
assigns execution roles.
\textbf{(c) PointFlow}: Multi-view observations are geometrically aligned and used by
Point DiT to jointly predict relative 3D point flow and gripper states.
Predicted object motion is converted to end-effector trajectories
through a learning-free geometry adapter for closed-loop execution.
}
    \label{fig:method}
    \vspace{-10pt}
\end{figure}




\subsection{Preliminary: Task-focused objects for 3D Point Flow Learning}
\label{sec:carrier_space}
\vspace{-5pt}
We define a \emph{task-focused object} as a physical entity whose motion characterizes the manipulation required at the current stage and can be mapped to the motion of one or both grippers. Specifically, for each gripper, the task-focused object is either the gripper itself or a locally rigid object part currently coupled to it. To support different manipulation modes, each task-focused object is additionally assigned an \emph{execution role}, indicating the gripper to which its predicted point flow should be mapped.

For example, when the left gripper approaches an object, the left gripper itself serves as the task-focused object with the execution role \emph{Left}. Once the object is grasped and transported, the task-focused object switches from the gripper to the coupled object, whose motion then directly characterizes the desired manipulation.

Under this formulation, PointPilot represents both gripper motion and manipulated-object motion within a shared point-flow space by casting manipulation as the motion of task-focused objects. 
\vspace{-5pt}
\subsection{Subtask Planning}
\vspace{-5pt}
\label{sec:architecture}
We use a memory-conditioned VLM planner to identify the current
manipulation stage and anticipate the next subtask. At planning
time $t$, the planner receives the main-view RGB image
$I_t^{\mathrm{main}}$, the global language instruction $\ell$,
and a textual memory bank $\mathcal{M}_t$ that records previously
predicted subtasks. At the beginning of each episode,
the memory bank is initialized to \texttt{NONE}.

Using Qwen2.5-VL as the backbone, the planner generates a
structured plan containing \texttt{CURRENT} and \texttt{NEXT} spans:
\begin{equation}
    (\hat{s}_t,\hat{s}_t^+)
    =
    VLM_{\theta}
    \left(I_t^{\mathrm{main}},\ell,\mathcal{M}_t\right).
    \label{eq:subtask_planning}
\end{equation}
Each span specifies a left/right subtask pair, with \texttt{NONE}
indicating an inactive hand. The \texttt{CURRENT} span
$\hat{s}_t$ describes the subtasks to execute at the current
stage. The \texttt{NEXT} span $\hat{s}_t^+$ specifies the
immediate next distinct subtask pair expected to begin within
the $H$-step horizon, or \texttt{NONE} if no new next subtask
is expected. The hidden states associated with
\texttt{CURRENT} provide task features
$\mathbf{Z}_t^{\mathrm{task}}$ for subsequent 3D object
grounding and point-flow prediction.

For the next planning call at $t+H$, the memory is updated
using only the previously predicted subtask spans:\begin{equation}
    \mathcal{M}_{t+H}
    =
    \operatorname{Concat}
    \left(\mathcal{M}_t,\hat{s}_t,\hat{s}_t^+\right).
    \label{eq:memory_update}
\end{equation}
The detailed design of VLM prompt and memory bank is indicated in the Appendix~\ref{app:method}.
\subsection{language-guided 3D Grounding}
\label{sec:grounding}
\vspace{-5pt}
\paragraph{Geometry Frustum} 
Inspired by PETR \citep{liu2022petrpositionembeddingtransformation, liu2022petrv2unifiedframework3d} in autonomous driving and DexBEV \citep{zhou2026dexteritybevaligning3dworld} in embodied intelligence, we construct the discretized camera frustums for frames. Specifically, we first inject camera-aware 3D positional information into the main-view image tokens. For an image token at
$\tilde{\mathbf{u}}_i=[u_i,v_i,1]^\top$, we sample $D$ depth
hypotheses along its corresponding camera ray using linearly increasing
discretization (LID).
Let $N_{\mathrm{img}}$ denote the number of main-view image tokens
and $D$ the number of depth hypotheses. For
$i\in\{1,\ldots,N_{\mathrm{img}}\}$ and
$k\in\{0,\ldots,D-1\}$, we define
\begin{equation}
z_k
=
z_{\min}
+
\frac{z_{\max}-z_{\min}}{D(D+1)}k(k+1)
\in [z_{\min},z_{\max}),
\qquad
\mathbf{x}_{i,k}^{c}
=
z_k\tilde{\mathbf K}^{-1}\tilde{\mathbf u}_i
\in\mathbb R^3 .
\label{eq:ray_backprojection}
\end{equation}
where
$\tilde{\mathbf K}\in\mathbb{R}^{3\times3}$ is the main-camera intrinsic matrix, and $c$ denotes the
main-camera coordinate frame. Following these equations, we obtain a ray-aligned 3D coordinate tensor
$\mathbf X_i^c\in\mathbb R^{D\times3}$ for each main-view image token.
Collectively, these coordinates form a discretized camera frustum
$\mathbf X^c\in\mathbb R^{N_{\mathrm{img}}\times D\times3}$. 

The resulting camera-frame coordinates are then encoded into
a 3D positional embedding $\mathbf e_i^{\mathrm{3D}}\in\mathbb R^C$, which is added to the corresponding final-layer
VLM image feature $\mathbf H_{t,i}^{\mathrm{img}}$, obtaining
$d$-dimensional geometry-aware image tokens.
This construction requires neither observed nor predicted depth.
\vspace{-5pt}
\paragraph{Grounding \& Classification} 
We concatenate these geometry-aware image tokens with the planner's
current-subtask token features $\mathbf{Z}_t^{\mathrm{task}}$ to form the grounding memory.
A six-layer transformer decoder updates $Q$ learnable queries through self-attention and cross-attention to this memory.
Each decoded query predicts a 3D box with center
$\mathbf c\in\mathbb R^3$, dimensions
$\boldsymbol\eta\in\mathbb R^3$, and a 6D rotation
representation $\mathbf r\in\mathbb R^6$,
along with an execution role from
$\{\mathrm{invalid},\mathrm{Left},\mathrm{Right},
\mathrm{Left\&Right}\}$.

The execution role associates each task-focused object
with the gripper(s) responsible for its manipulation:
$\mathrm{Left}$ routes this object's point flow to the left gripper, and the $\mathrm{Left\&Right}$ denotes that both grippers are coupled to the same object and follow its motion.

During training, we use Hungarian matching to establish
a one-to-one assignment between predicted queries and
ground-truth task-focused objects.
Matched queries are supervised with their assigned
3D boxes and execution roles, while unmatched queries
are assigned the $\mathrm{invalid}$ role and excluded
from box regression.
The matching cost and loss definitions are detailed
in the Appendix~\ref{app:method}.

\subsection{Point Flow DIT}
\vspace{-5pt}

\paragraph{Spatial Alignment}
We use the main-view coordinate as the anchored reference frame for both observations and point flow prediction. Auxiliary-view images are encoded by the shared vision encoder.
Following Sec.~\ref{sec:grounding}, we construct their discretized
camera frustums and transform the sampled 3D coordinates into
the main-camera frame using calibrated camera extrinsics:
\begin{equation}
    \tilde{\mathbf x}_{i,k}^{c}
    =
    \mathbf T_{c_v \rightarrow c}
    \tilde{\mathbf x}_{i,k}^{c_v},
    \qquad
    v\in\mathcal V_{\mathrm{aux}}
\end{equation}
where $c_v$ denotes the coordinate frame of auxiliary views $v$, and
$\mathbf T_{c_v \rightarrow c}$ is the coordinate
transformation from auxiliary views $v$ to the main view.
These aligned frustums provide 3D positional embeddings
for the auxiliary-view visual features.
\vspace{-5pt}
\paragraph{Point Flow Prediction}
For each grounded object, we use Halton sampling within its predicted 3D bounding box to obtain
$N$ initial points.
PointPilot predicts their future displacements
$\Delta\mathbf p_h=\mathbf p_h-\mathbf p_0$
over an $H$-step horizon.

 To augment the condition of the grounded object, each grounded object is represented by an object token $\mathbf Z_t^{\mathrm{obj}}$. This token fuses
geometry features of its sampled initial points processed by a light PointNet with grounded visual appearance features extracted by Region of Interest Align (ROIAlign) from the projected 3D bounding box region.

For point-flow generation, each future point displacement is represented
by a noisy point token. In order to preserve its spatial identity, we augment each
token with a spatial position embedding of the corresponding current 3D anchor point in the main-camera coordinate frame. Therefore, the model predicts motion from spatially grounded
point tokens rather than from pure noise. Furthermore, at each future step, noisy point tokens are interleaved with learned gripper tokens, enabling joint prediction of object motion and
gripper states. We train the DiT with conditional flow matching to predict the velocity field that transports noisy point tokens
to future point displacements. Gripper tokens are not noised and directly predict binary open/close states without the noising-denoising process.

The Point DiT is conditioned on the grounding memory, object tokens,
and spatially aligned auxiliary-view visual features. 


\vspace{-5pt}
\subsection{From Point Flow to Executable Robot Actions}
\vspace{-5pt}
Before grasping, predicted gripper flow guides execution.
After grasping, object flow guides the coupled gripper,
assuming a rigid grasp.
Our learning-free adapter transforms current and predicted
points into the world frame and fits a rigid motion:
\begin{equation}
(\hat{\mathbf R}_{t,h}^{m},\hat{\mathbf d}_{t,h}^{m})
=
\arg\min_{\mathbf R\in\mathrm{SO}(3),\,\mathbf d\in\mathbb R^3}
\sum_{n=1}^{N}
\left\|
\mathbf R\mathbf p_{t,n}^{m,w}+\mathbf d
-\hat{\mathbf p}_{t+h,n}^{m,w}
\right\|_2^2.
\end{equation}
The fitted relative transform $\Delta\hat{\mathbf T}$
updates each assigned end effector's current pose
$\mathbf T_a$ to its target pose
$\hat{\mathbf T}_a=\Delta\hat{\mathbf T}\mathbf T_a$,
with all poses expressed in the world frame.
For $\mathrm{Left\&Right}$, this transform acts on both
grippers' current poses; independent motions use
separate object flows.
The resulting poses and predicted gripper states are
executed through IK and the robot controller.
PointPilot observes the updated scene and replans after
each action chunk.
\vspace{-5pt}
\subsection{Training Objective}
\vspace{-5pt}
\label{sec:learning}
PointPilot is jointly optimized for subtask planning,
3D grounding, point-flow prediction, and gripper control:
\begin{equation}
    \mathcal L
    =
    \lambda_{1}\mathcal L_{\mathrm{plan}}
    +
    \lambda_{2}\mathcal L_{\mathrm{ground}}
    +
    \lambda_{3}\mathcal L_{\mathrm{pointflow}}
    +
    \lambda_{4}\mathcal L_{\mathrm{gripper}}.
\end{equation}
The VLM is trained with cross entropy in a teacher forcing way.
Grounding combines 3D box regression and execution-role
classification under Hungarian matching.
Conditioned on predicted bounding boxes,
point-flow prediction uses conditional flow matching
on anchor-relative displacements, and gripper states
are supervised with binary cross entropy.
Full loss definitions, matching costs, masking rules,
and optimization settings are provided in the Appendix~\ref{app:method}.

\par\vspace{-5pt}
\section{Data Pipeline for Task-focused Objects Point Flow}
\label{sec:data_construction}
\vspace{-5pt}

We construct PointPilot's stage-level supervision from simulated and
real-robot demonstrations through interaction-stage segmentation and
3D trajectory extraction. This aligns task semantics with task-focused
object motion without requiring shared low-level action encodings across data sources.
\vspace{-5pt}
\subsection{Stage Segmentation}
\label{sec:data_stage_segmentation}
\vspace{-5pt}
We primarily segment demonstrations using gripper open/close events.
In simulation, gripper states and grasp relationships define interaction
boundaries; for real-robot data, the same events are detected from the
smoothed gripper-width signal. Contact and articulated-part motion cues are
used when gripper events are insufficient. Bimanual interactions are aligned
on a shared timeline, and offline stage-level language annotations are matched
to the resulting segments.
\vspace{-5pt}
\subsection{Trajectory Extraction}
\label{sec:data_trajectory_extraction}
\vspace{-5pt}
\paragraph{Simulation.}
We directly retrieve EEF and object pose trajectories from simulator states.
For articulated interactions, we track the manipulated moving part rather
than the full object assembly. All trajectories are transformed into a
consistent reference frame and aligned with the segmented stages.
\vspace{-20pt}
\paragraph{Real-world.}
EEF trajectories are obtained directly from recorded robot poses. For
grasped-object stages, we recover object geometry once at a visual anchor and
propagate it using the recorded EEF motion, avoiding frame-wise tracking.
Following the visual-grounding pipeline of
MolmoMotion~\citep{zhang2026molmomotion}, we combine MolmoPoint
localization~\citep{clark2026molmopoint} with SAM~3 masks
\citep{carion2025sam3}, and lift the selected region to 3D using estimated
depth and camera calibration.

Let $T_B^A$ denote the transform from frame $B$ to frame $A$, and let $a$
be the anchor time. Assuming rigid object--gripper coupling, the object pose
during the grasp interval is reconstructed as
\begin{equation}
\label{eq:eef_driven_object_reconstruction}
\widehat T^C_{O_t}
=
T^C_{E_t}
\left(T^C_{E_a}\right)^{-1}
\widehat T^C_{O_a},
\qquad
t\in\mathcal I_{\mathrm{coupled}} .
\end{equation}
Here, $\widehat T^C_{O_a}$ is the visually estimated anchor pose, while
recorded EEF motion provides its temporal propagation. Frames outside the
accepted coupling interval are not extrapolated.


Appendix~\ref{app:data_construction} details validity checks and failure handling for depth calibration,
anchor selection, mask filtering, and reconstruction. We align stage
intervals, language annotations, and valid trajectories by their original
timestamps to construct sparse point-flow supervision.
\par\vspace{-5pt}
\section{Experiments}
\vspace{-5pt}
We evaluate PointPilot from three perspectives:
(1) general manipulation performance on single-arm and bimanual simulation benchmarks and real-world tasks,
(2) zero-shot skill transfer across previously unseen robot
embodiments, and (3) the controlled modular ablations to indicate the effectiveness of each module especially the necessity of spatial alignment.
\vspace{-5pt}
\subsection{Simulation Result}
\label{sec:exp_manipulation}
\vspace{-5pt}
\paragraph{Training and evaluation setup.}
PointPilot is jointly trained on demonstrations from
LIBERO and RoboTwin, combining different robot embodiments
and single-arm and bimanual manipulation within one model.
The results on both benchmarks therefore evaluate a shared
policy learned from heterogeneous demonstrations.
We follow the standard evaluation protocols of LIBERO
and RoboTwin 2.0 Clean and report task success rates.
\vspace{-5pt}
\paragraph{Single-arm manipulation on LIBERO.}
As shown in Table~\ref{tab:libero}, PointPilot achieves 85.8\% average success across the four LIBERO suites,
including the best listed results on LIBERO-Spatial (96.0\%) and
LIBERO-Object (97.6\%). It obtains 77.2\% on LIBERO-Long. Performance on
LIBERO-Goal is mainly limited by three non-coupled, friction-driven tasks;
excluding them, success reaches 96.0\%. These results demonstrate competitive single-arm
manipulation using object-centric point flow without
a learned end-effector action decoder.
\vspace{-5pt}
\begin{table}[!htbp]
\centering
\vspace{5pt}
\begin{minipage}[t]{0.58\linewidth}
\vspace{0pt}
\centering
\setlength{\abovecaptionskip}{0pt}
\setlength{\belowcaptionskip}{2pt}
\captionof{table}{Success rates (\%) on LIBERO. All baseline results follow the evaluation reported by
RoboFlow4D~\citep{lin2026roboflow4d}.}
\label{tab:libero}

\scriptsize
\setlength{\tabcolsep}{2.0pt}
\renewcommand{\arraystretch}{1.04}

\begin{adjustbox}{max width=\linewidth}
\begin{tabular}{@{}lccccc@{}}
\toprule
Method & Spatial & Object & Goal & Long & Average \\
\midrule

Octo~\citep{ghosh2024octo}
& 78.9 & 85.7 & 84.6 & 51.1 & 75.1 \\

CogACT~\citep{li2024cogact}
& 87.5 & 90.2 & 78.4 & 53.2 & 77.3 \\

OpenVLA~\citep{kim2025openvla}
& 84.7 & 88.4 & 79.2 & 53.7 & 76.5 \\

TraceVLA~\citep{zheng2024tracevla}
& 84.6 & 85.2 & 75.1 & 54.1 & 74.8 \\

SpatialVLA~\citep{qu2025spatialvla}
& 88.2 & 89.9 & 78.6 & 55.5 & 78.1 \\

4D-VLA~\citep{zhang2025vla}
& 88.9 & 95.2 & 90.9 & \textbf{79.1} & 88.6 \\

DP w/ RoboFlow4D~\citep{lin2026roboflow4d}
& 89.8 & 93.2 & 85.2 & 72.0 & 85.1 \\

DiT w/ RoboFlow4D~\citep{lin2026roboflow4d}
& 90.2 & 97.0 & 88.4 & 75.2 & 87.7 \\

UniAct (ACT head)~\citep{zheng2025uniact}
& 77.0 & 87.0 & 77.0 & 70.0 & 77.8 \\

SmolVLA (2.25B)~\citep{shukor2025smolvla}
& 93.0 & 94.0 & \textbf{91.0} & 77.0 & \textbf{88.8} \\

\midrule
\textbf{PointPilot (Ours)}
& \textbf{96.0}
& \textbf{97.6}
& 72.4
& 77.2
& 85.8 \\
\bottomrule
\end{tabular}
\end{adjustbox}


\end{minipage}
\hfill
\begin{minipage}[t]{0.39\linewidth}
\vspace{0pt}
\centering
\setlength{\abovecaptionskip}{0pt}
\setlength{\belowcaptionskip}{2pt}
\captionof{table}{Success rates (\%) on RoboTwin 2.0. Baseline results are reported by RoboTwin 2.0. The GR00T-N1.7 result is from the Easy setting of RoboTwin-Clean2Rand in Qwen-RobotManip~\citep{qwenrobotmanip2026}.}
\label{tab:robotwin}

\scriptsize
\setlength{\tabcolsep}{2.5pt}
\renewcommand{\arraystretch}{1.04}

\begin{adjustbox}{max width=\linewidth}
\begin{tabular}{@{}lc@{}}
\toprule
Method & Success rate (\%) \\
\midrule

DP
& 28.0 \\

ACT
& 29.7 \\

RDT-1B
& 34.5 \\

GR00T-N1.7
& 43.6 \\

$\pi_0$
& 46.4 \\

DP3
& 55.2 \\

\midrule
\textbf{PointPilot (Ours)}
& \textbf{56.6} \\

\bottomrule
\end{tabular}
\end{adjustbox}


\end{minipage}
\vspace{-15pt}
\end{table}

\paragraph{Bimanual manipulation on RoboTwin.}
The same checkpoint achieves 56.6\% success
on RoboTwin 2.0 Clean, compared with the reported 55.2\%
of DP3.
Together with the LIBERO results, this demonstrates
that a shared object-centric policy can learn from
heterogeneous demonstrations and support both
single-arm and bimanual manipulation.
\vspace{-5pt}
\subsection{Zero-Shot Cross-Embodiment Transfer}
\label{sec:exp_transfer}
\vspace{-5pt}
We evaluate zero-shot transfer from the ALOHA embodiment
in RoboTwin to ARX and Piper on the tasks in
Table~\ref{tab:cross_embodiment}.
The jointly trained policy is deployed without
target-embodiment demonstrations or fine-tuning.
Predicted object motion is converted into executable
commands through the geometric adapter and each
target robot's kinematics and controller.

\begin{table}[!htbp]
\centering
\setlength{\abovecaptionskip}{0pt}
\setlength{\belowcaptionskip}{3pt}
\caption{Cross-embodiment success rates (\%) on eight tasks (100 trials per task). For each embodiment, the experiment seeds are the same. X-VLA is tested with its official checkpoint not trained on ARX and Piper. In robotwin, not all tasks are valid in all embodiments. ``--'' indicates this task is incompatible with this embodiment. $^{*}$ denotes the average and $\Delta$ are calculated across 7 tasks------the alarm-clock task is excluded.  }
\label{tab:cross_embodiment}
\scriptsize  
\setlength{\tabcolsep}{2pt}
\renewcommand{\arraystretch}{1.12}
\begin{tabular*}{\linewidth}{@{\extracolsep{\fill}}lrrrrrr@{}}
\toprule
Task & \multicolumn{2}{c}{ALOHA}
     & \multicolumn{2}{c}{ARX}
     & \multicolumn{2}{c}{Piper} \\
\cmidrule(lr){2-3}\cmidrule(lr){4-5}\cmidrule(l){6-7}
 & PointPilot & X-VLA & PointPilot & X-VLA & PointPilot & X-VLA \\
\midrule
Grab roller               & 98.0 & 100.0 & 85.0 & 79.0  & 79.0 & 21.0 \\
Shake bottle              & 100.0 & 100.0 & 93.0 & 90.0  & 75.0 & 41.0 \\
Shake bottle horizontally & 94.0 & 99.0 & 90.0 & 90.0  & 83.0 & 41.0 \\
Click alarm clock         & 75.0  & 17.0   & 69.0 & 62.0  & --  & -- \\
Click bell                & 79.0  & 99.0 & 75.0 & 97.0 & 77.0  & 1.0 \\
Dump bin into big bin     & 100.0 & 81.0 & 67.0 & 29.0  & 69.0  & 24.0 \\
Stack two bowls           & 54.0  & 62.0  & 41.0 & 16.0  & 39.0  & 0.0 \\
Pick dual bottles         & 88.0  & 89.0 & 85.0 & 93.0  & 75.0  & 49.0 \\
\midrule
\textbf{Average}
& 86.0 & 80.9 & 75.6 & 69.5 & 71.0$^{*}$ & 25.3$^{*}$ \\

$\textbf{Delta}$ \textbf{compared with ALOHA (pp)}
& -- & -- & \textbf{-10.4} & -11.4 & \textbf{-16.6$^{*}$} & -64.7$^{*}$ \\
\bottomrule
\end{tabular*}
\par\vspace{-10pt}
\end{table}

As shown in Table ~\ref{tab:cross_embodiment}, PointPilot presents a much smaller transfer gap, especially in Piper, compared with X-VLA. Piper differs more visibly from ALOHA than ARX does.
Thus, X-VLA exhibits a substantially larger transfer gap on Piper, whereas PointPilot maintains comparable performance across the two target embodiments, suggesting that its object-centric
3D motion representation reduces sensitivity to robot appearance.

\subsection{Real World Results}
\begin{figure}[!htbp]
\centering

\sbox{\realworldbox}{%
    \includegraphics[width=0.81\linewidth]{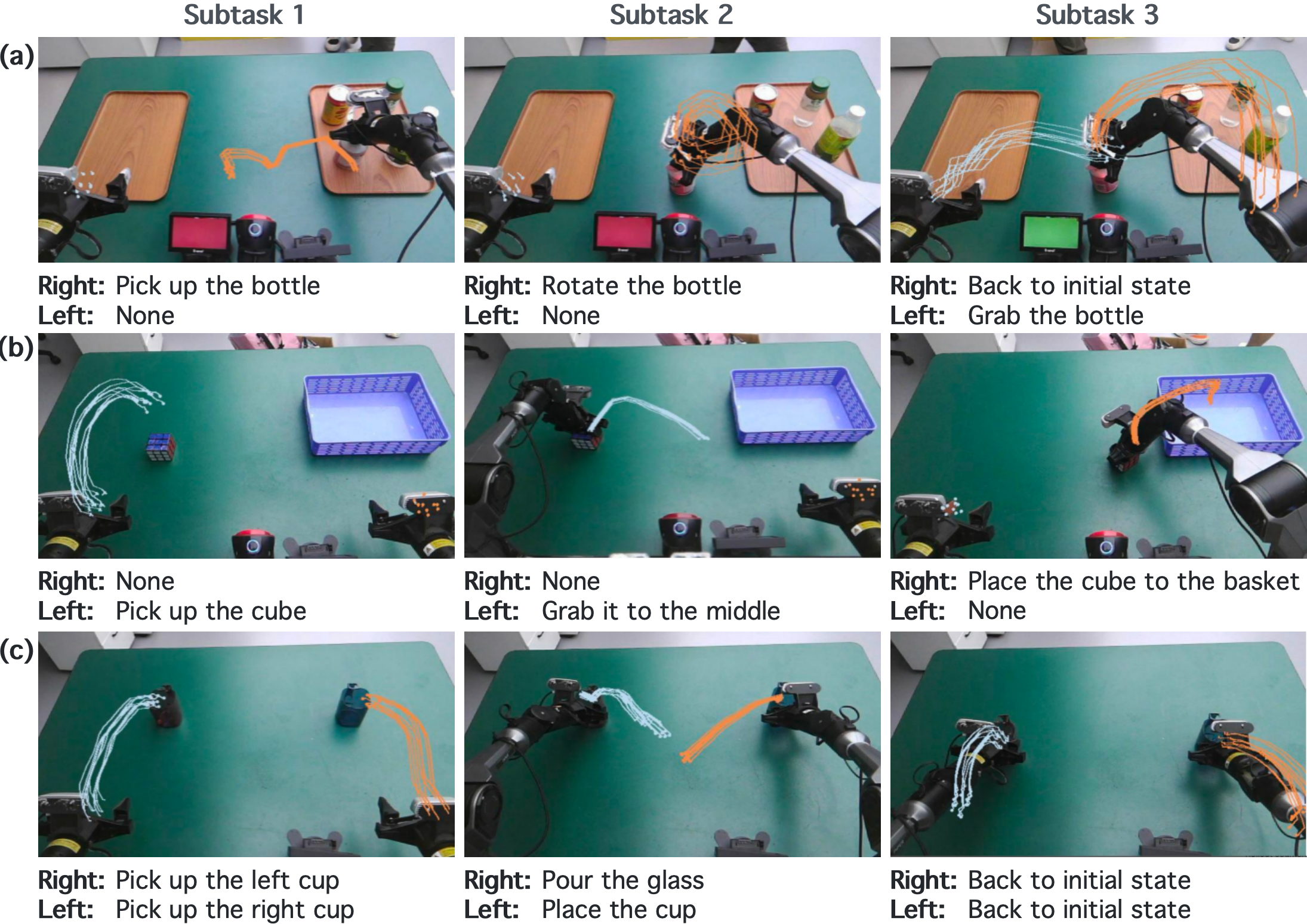}%
}
\setlength{\realworldheight}{%
    \dimexpr\ht\realworldbox+\dp\realworldbox\relax
}
\setlength{\realworldcolwidth}{0.18\linewidth}

\newcommand{\rwresult}[4]{%
    \vbox to #1\realworldheight{%
        \vbox to 0.245\realworldheight{%
            \vfil
            \hbox{%
                \sffamily\fontsize{7}{8.5}\selectfont
                \setlength{\tabcolsep}{0pt}%
                \setlength{\abovetopsep}{0pt}%
                \renewcommand{\arraystretch}{1.05}%
                \begin{tabular*}{\realworldcolwidth}
                    {@{\extracolsep{\fill}}lr@{}}
                    \toprule
                    \multicolumn{2}{@{}l@{}}{%
                        \parbox[t]{\realworldcolwidth}{%
                            \raggedright\bfseries
                            \strut #2\strut
                        }%
                    }\\[3pt]
                    Trials    & 20 \\
                    Successes & #3 \\
                    Rate (\%) & \textbf{#4} \\
                    \bottomrule
                \end{tabular*}%
            }%
            \vfil
        }%
        \vfil
    }%
}

\usebox{\realworldbox}\hspace{0.01\linewidth}%
\vbox to \realworldheight{%
    \offinterlineskip
    \vbox to 0.04\realworldheight{%
        \kern0.004\realworldheight
        \hbox to \realworldcolwidth{%
            \hfil
            {\sffamily\fontsize{8}{9.5}\selectfont
             \bfseries Results}%
            \hfil
        }%
        \vfil
    }%
    \rwresult{0.321}
        {Retail QR code\\scanning}{10}{50}%
    \rwresult{0.323}
        {Placing a Rubik's\\cube in a bin}{18}{90}%
    \rwresult{0.316}
        {Pouring from a cup\\to another cup}{14}{70}%
    \vfil
}
\par\vspace{-3mm}
\caption{Autonomous closed-loop real-world manipulation with PointPilot.
Each row shows successive subtasks and predicted point flows.
The rightmost column reports successful trials and success rates
over 20 trials per task, with an overall success rate of 70\%.}
\label{fig:realworld}
\end{figure}

\vspace{-5pt}
We evaluate PointPilot on three real-world tasks: retail QR code scanning, 
placing a Rubik's cube into a bin, and pouring from a cup, with 20 trials
per task. The policy operates autonomously in a closed loop, replanning from
updated observations after each action chunk.
Figure~\ref{fig:realworld} reports success rates of 50\% (10/20), 90\% (18/20) and 70\% (14/20) respectively, yielding an overall success rate
of 70\% (42/60).

The rollouts illustrate distinct demands on task-focused object motion:
reorienting a bottle for scanning, sequencing per-arm subtasks for cube placement,
and positioning and tilting containers for pouring. These examples
connect subtask planning and point-flow prediction to physical execution
involving both translation and rotation. Cube placement has the highest
observed success rate, while scanning leaves the largest room for
improvement. Together, the results demonstrate the feasibility of the
task-focused point-flow interface beyond simulation, while performance
remains task-dependent. Rollout videos are provided in the \textbf{supplementary material}.

\vspace{-5pt}
\subsection{Ablation Studies}
\vspace{-5pt}
\begin{figure}[!htbp]
    \centering

    \begin{minipage}[t]{0.42\linewidth}
        \vspace{0pt}
        \centering
        \setlength{\abovecaptionskip}{0pt}
        \setlength{\belowcaptionskip}{2pt}

        \captionof{table}{
        Effect of the planning memory bank on LIBERO success rates. Memory consistently improves performance across all four suites, with the largest gain on LIBERO-Long (+22.2 points). Without planning history, the VLM frequently repeats an already executed
subtask during consecutive replanning steps, causing the policy to stall.
        }
        \vspace{-5pt}
        \label{tab:memory_ablation}
        \vspace{3mm}
        \footnotesize
        \setlength{\tabcolsep}{3.2pt}
        \renewcommand{\arraystretch}{1.08}

        \begin{tabular}{@{}lccc@{}}
            \toprule
            Suite & w/o Memory & PointPilot & $\Delta$ \\
            \midrule
            Spatial & 90.4 & \textbf{96.0} & \textbf{+5.6} \\
            Object  & 89.6 & \textbf{97.6} & \textbf{+8.0} \\
            Goal    & 65.8 & \textbf{72.4} & \textbf{+6.6} \\
            Long    & 55.0 & \textbf{77.2} & \textbf{+22.2} \\
            \midrule
            Average & 75.2 & \textbf{85.8} & \textbf{+10.6} \\
            \bottomrule
        \end{tabular}

    \end{minipage}
    \hfill
    \begin{minipage}[t]{0.56\linewidth}
        \vspace{0pt}
        \centering
        \setlength{\abovecaptionskip}{0pt}

        \includegraphics[
            width=0.95\linewidth
        ]{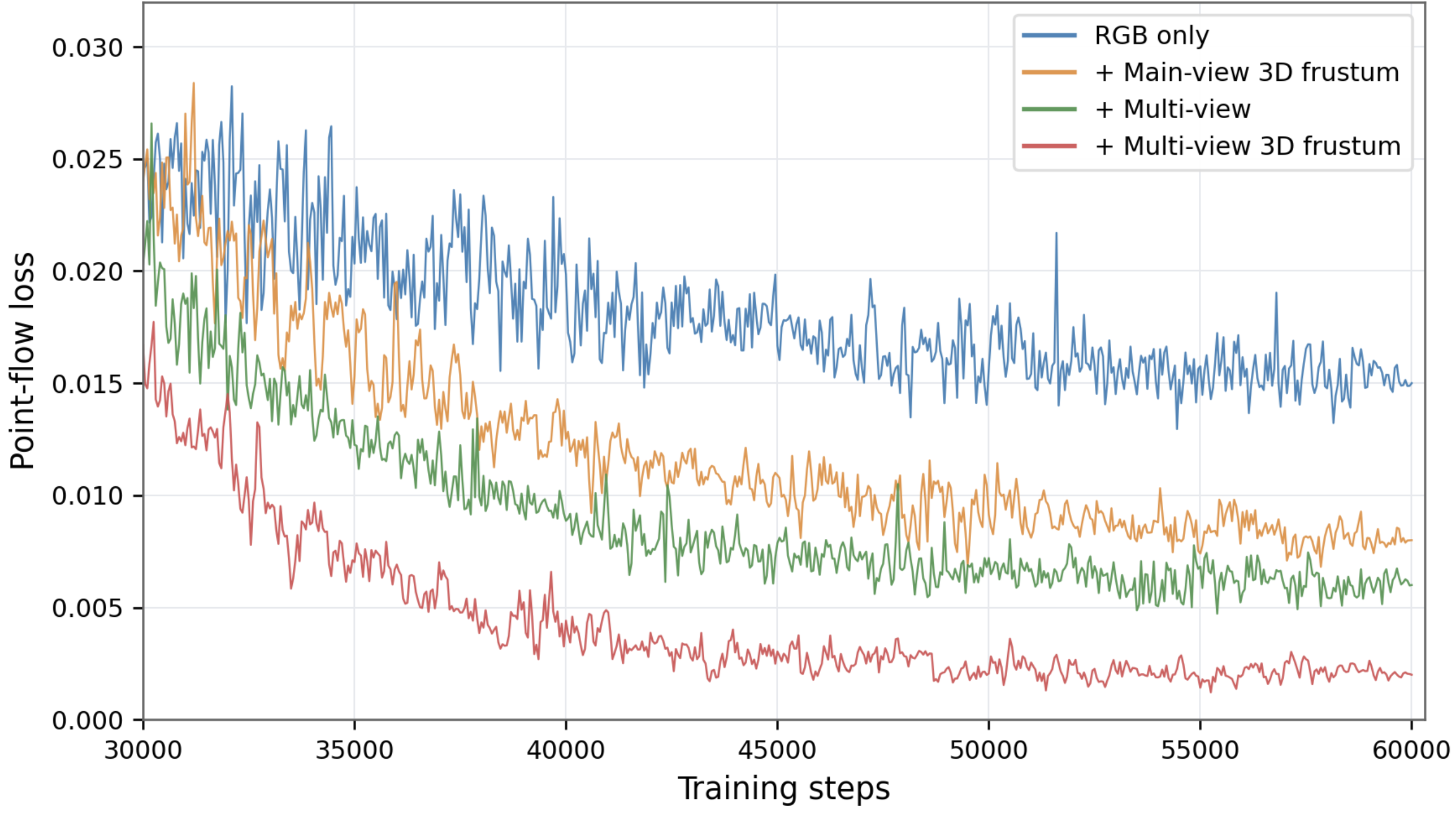}
        \captionof{figure}{
        Point-flow loss under different visual-frustum configurations when trained jointly on LIBERO and RoboTwin.
        Main-view 3D frustum encoding, additional views and only main-view 3D frustum encoding, and multi-view 3D frustum alignment progressively improve convergence.
        }
        \label{fig:geometry_ablation}

    \end{minipage}

\end{figure}
\vspace{-5pt}
We ablate the planning memory bank and 3D geometric alignment, measuring
their effects on LIBERO task success and point-flow learning, respectively.
\vspace{-5pt}
\paragraph{Effect of planning memory.}
As shown in Table~\ref{tab:memory_ablation}, removing planning memory lowers
average success from 85.8\% to 75.2\%, with the largest drop on LIBERO-Long
(77.2\% to 55.0\%). The 22.2 percentage-point gap on Long is larger than the
5.6 --8.0 point gaps on the other suites, suggesting that planning history is
especially useful for longer subtask sequences. Without history, the VLM can
repeat already executed subtasks during replanning and stall execution.
Retaining prior plan predictions therefore complements the current visual
input with temporal context for subtask selection and stage progression.
\vspace{-5pt}
\paragraph{Effect of 3D frustum and multi-view alignment.}
Figure~\ref{fig:geometry_ablation} compares visual-geometric configurations
under joint LIBERO and RoboTwin training. Adding main-view 3D frustum
encoding reduces the final point-flow loss from 0.015 to 0.008, and auxiliary
views further reduce it to 0.006. Applying spatially aligned 3D frustum
encoding to all views reaches 0.002. In particular, loss decreases further
after adding aligned geometric encoding even when multi-view images are
already available. This suggests that additional views and geometric
alignment make complementary contributions. The results support expressing
visual geometry in the same coordinate frame as predicted point motion,
rather than relying solely on multi-view appearance.
\par\vspace{-5pt}
\section{Conclusion and Future Work}
\label{sec:conclusion}
\vspace{-5pt}

PointPilot represents subtask-focused objects with sparse 3D points and execution roles, providing a shared interface across supported manipulation modes and embodiments. Memory-conditioned subtask planning and grounding connect task intent to point-flow prediction, with multi-view geometry and predicted motion aligned in the main-view coordinate. Under stage-local rigid coupling, a non-learned geometric adapter converts predicted flows into end-effector motion without an additional learned continuous-action decoder.

Several limitations suggest directions for further research. The current
formulation centers on locally rigid task-focused objects and gripper-based
execution. As a scalable object-centric framework, we need to leverage more data (including human hands and dexterous hands) to prove the scalability and show the generalization. Moreover, prediction focuses on the
task-focused objects selected for the current stage rather than explicitly
forecasting the broader set of task-relevant entities. Jointly modeling
manipulated objects and other entities affected by their motion could support
richer reasoning about object--object interactions and indirect effects while
retaining a sparse, task-focused representation. Finally, reducing the burden of aligning stage semantics, object geometry, execution roles, and motion while preserving semantic–geometric consistency remains a challenge for scaling.

\bibliography{iclr2027_conference}
\bibliographystyle{iclr2027_conference}

\clearpage
\appendix

\section{Data Construction Details}
\label{app:data_construction}

This appendix expands the interaction-stage segmentation and 3D trajectory
extraction used to construct task-focused point-flow supervision for PointPilot.
We briefly summarize simulation processing and focus on
gripper-event detection and geometric reconstruction for our self-collected
real-robot recordings. The output consists of stage-aligned entity poses and
geometry, together with their temporal correspondences, validity information, and
reconstruction status.

\subsection{Stage Segmentation}
\label{app:data_stage_segmentation}

Gripper opening and closing events provide the primary temporal anchors for
segmentation. Simulation supplies recorded gripper-state changes; for our
real-robot recordings, we recover the corresponding events from the continuous
gripper-width signal. We first apply median smoothing and compute the signed
numerical gradient with respect to the recorded timestamps. Let $s_i$ denote this
gradient at sample $i$. The automatic event threshold is

\begin{equation}
\label{eq:app_gripper_slope_threshold}
\tau_s=\max\!\left(\tau_{\min},Q_{0.9}(\{|s_i|\})\right),
\end{equation}

where $Q_{0.9}$ denotes the 90th percentile. Contiguous intervals exceeding this
magnitude threshold and satisfying a minimum duration in source frames form
candidate events. Their direction is determined by the median signed slope:
negative for closing and positive for opening.

Nearby events with the same direction are merged. A candidate is accepted only if
it agrees with the current open/closed state, reaches the opposite-state plateau,
and produces a sufficient change in width. Accepted closing events place a
boundary at the end of closure, whereas accepted opening events place one at the
onset of opening. The resulting intervals indicate candidate grasp states rather
than independently certifying object attachment. Bimanual intervals are
synchronized while retaining per-arm information. Simulation additionally uses
contact and moving-part cues for interactions not adequately captured by gripper
events, such as drawer manipulation. Stage-level language annotations are obtained
through a separate offline procedure and associated with the resulting intervals.

\subsection{Trajectory Extraction}
\label{app:data_trajectory_extraction}

For simulated demonstrations, we associate recorded object, moving-part, and
end-effector (EEF) poses with the corresponding entities and stage intervals, as
described in the main paper. Each pose sequence is paired with its entity geometry
in a consistent reference frame. We focus below on real-robot reconstruction,
where object trajectories must be recovered from visual observations and robot
motion rather than directly read from simulator states.

\paragraph{Depth Calibration and Camera Geometry.}

For our self-collected real-robot recordings, we estimate depth with
ViPE~\citep{huang2025vipe} on sparse frames associated with grasped-object stages.
These depth estimates are used for offline geometric reconstruction, not as
inputs to PointPilot at inference. Reconstruction uses the supplied fixed-camera
intrinsics and robot-to-camera transforms rather than ViPE's estimated camera
parameters. Estimated depth can
have substantial scale and offset errors relative to the robot reference. Even
with a correct object mask, directly back-projecting this depth can misplace the
object relative to the gripper and distort its reconstructed size. These errors
affect the object-to-EEF transform used to propagate object motion. We therefore
calibrate the estimated depth against recorded EEF positions before
reconstructing the visual anchor.

To construct calibration references, we use the supplied camera calibration to
express recorded EEF positions in the camera coordinate frame. Their optical-axis
coordinates $z_i^E$ provide the robot-derived depth references. Using the camera
intrinsics, we project each reference point into the corresponding depth image at
$(u_i,v_i)$. We then compute $d_i$, the median of finite, positive estimated depths
in a $15\times15$ pixel patch centered on this projection. This gives a pair
$(d_i,z_i^E)$ linking an image-based depth estimate to the robot reference. Usable
pairs from both arms and multiple sampled times are pooled to estimate a shared
correction.

We model the systematic discrepancy as $z_i^E\approx\alpha d_i+\beta$, where
$\alpha>0$ adjusts depth scale and $\beta$ adjusts offset. RANSAC identifies a
positive-scale relation supported by multiple observations and rejects pairs
with large residuals. Least-squares fitting on the current inliers then refines
the parameters, with the inlier set updated from the residuals. The resulting
correction is applied to the corresponding depth maps:

\begin{equation}
\label{eq:app_eef_depth_calibration}
D_{\mathrm{cal}}(u,v)=\alpha D_{\mathrm{est}}(u,v)+\beta.
\end{equation}

The same fitted parameters are applied to all associated frames, rather than
fitting each pixel or frame independently. Although the references are sampled
near the EEF projections, the correction is applied to the depth maps used for
object reconstruction, not only to those local patches.

We accept a fit only when it has sufficient supporting observations, adequate
depth variation and correlation, and a scale within the allowed range. If these
references are insufficient for a supported fit, the designated failure path
retains the supplied camera geometry and uncalibrated depth with an explicit
calibration-failure status. The EEF frame origin need not lie on a visible
gripper surface, so the sampled patch provides an approximate correspondence
rather than an exact depth measurement of that point. The calibration therefore
constrains local consistency with the robot reference without establishing
scene-wide depth ground truth or an object-trajectory error bound.

\paragraph{Visual Anchor Selection.}

For a grasped-object stage, we search for a visual anchor within the candidate
grasp interval. Adapting the visual-grounding components of
MolmoMotion~\citep{zhang2026molmomotion}, we use
MolmoPoint~\citep{clark2026molmopoint} to obtain image-plane location cues and
SAM~3~\citep{carion2025sam3} to generate candidate masks. Localization specifies
the manipulated object and the corresponding arm. MolmoPoint processes each
candidate image as a short repeated-frame clip; its locations guide selection from
text-conditioned SAM 3 masks. When no suitable text-conditioned mask is matched, a
small box around a predicted location provides an additional segmentation prompt.
These components localize candidate anchors rather than track the object
throughout the stage.

We evaluate up to five candidate frames and separate acceptance from ranking. A
supported candidate must meet foreground-area and mask-to-EEF proximity
thresholds. A stage requires at least $\min(3,N)$ supported candidates, where $N$
is the number examined. Among supported candidates, we select the highest-scoring
anchor using

\begin{equation}
\label{eq:app_anchor_selection_score}
Q=0.25Q_{\mathrm{area}}+0.40Q_{\mathrm{depth}}
  +0.15Q_{\mathrm{boundary}}+0.20Q_{\mathrm{proximity}},
\end{equation}

where the terms quantify mask area, valid-depth coverage, image-boundary margin,
and proximity to the projected EEF of the corresponding arm. Support across
candidate frames is used as a consistency check; only the selected anchor is used
to recover object geometry. We do not fuse multiple anchor poses or interpret the
score as a calibrated probability of reconstruction accuracy.

\paragraph{Object Geometry and Motion Reconstruction.}

Following the reconstruction convention in the main paper, we express
reconstructed object geometry in the supplied fixed-camera frame $C$. At the
selected anchor time $a$, valid masked depth pixels are back-projected using the
supplied intrinsics:

\begin{equation}
\label{eq:app_anchor_backprojection}
\mathbf p^C_a(u,v)=D_a(u,v)
\begin{bmatrix}
(u-c_x)/f_x\\
(v-c_y)/f_y\\
1
\end{bmatrix},
\end{equation}

where $D_a$ is the accepted depth map, using the affine correction when available.
We cap the point cloud at 30,000 samples, remove the most distant 10\% relative to
its coordinate-wise median, and compute a right-handed PCA frame. The 2nd and 98th
percentiles along each axis determine a robust box center and extent, with a
minimum extent for nearly degenerate dimensions. The resulting anchor pose
$\widehat T^C_{O_a}$ and oriented bounding box (OBB) approximate the visible
object region rather than a complete CAD-aligned object model.

We recover temporal object motion through the fixed relative transform described
in the main paper:

\begin{equation}
\label{eq:app_eef_object_propagation}
\widehat T^E_O=
\left(T^C_{E_a}\right)^{-1}\widehat T^C_{O_a},
\qquad
\widehat T^C_{O_t}=T^C_{E_t}\widehat T^E_O,
\quad t\in\mathcal I_{\mathrm{coupled}}.
\end{equation}

The same transformation propagates the object-local OBB vertices. Thus, visual
reconstruction estimates the anchor geometry once, while recorded EEF poses
provide the subsequent motion. This construction assumes approximately rigid
attachment without appreciable slip within $\mathcal I_{\mathrm{coupled}}$; the
existence of a stage alone does not establish that assumption. Outside the
reconstruction interval, including after release, object poses and vertices are
marked unavailable rather than propagated with the former gripper attachment.
Errors in anchor geometry, camera alignment, and the attachment assumption can
still affect the reconstructed trajectory.

\paragraph{Gripper Geometry and Fallback Handling.}

We construct a sequence of gripper 3D bounding box from URDF collision geometry, recorded joint
states, and EEF poses. The joint states determine the gripper configuration, while
the EEF poses place the geometry in the scene. Before combining these sources, we
check that the logged EEF poses are consistent with forward-kinematic flange poses
under the supplied reference-frame mapping. The resulting bounding boxes describe the
gripper's motion and opening without requiring image-based reconstruction.

When object localization or geometric reconstruction fails to produce an accepted
object trajectory, we fall back to the corresponding gripper trajectory and OBB
derived from recorded EEF poses and robot geometry. Fallback cases are explicitly
flagged and treated as gripper geometry rather than successfully reconstructed
object motion.

Finally, stage intervals, language annotations, and geometric outputs are
associated through source-frame indices and timestamps. Reconstruction validity
and fallback provenance are retained separately. The stored object masks
correspond to selected anchors rather than dense per-frame segmentation, and the
sparse geometry timeline is not assumed to be uniformly sampled. These records
provide pose sequences and validity information for constructing PointPilot's
point-flow supervision for task-focused objects. The prediction targets are
future 3D point displacements relative to their current positions, as defined in
the main paper.
\section{More Details about PointPilot}
\label{app:method}

To those who want to reproduce our work, we detail the computation about PointPilot in this section. At planning anchor $t$,
all boxes and point trajectories use the current main-camera frame $c_t$.
The world-to-camera extrinsic is $\mathbf E_t$; bars denote homogeneous
coordinates, with $\bar{\mathbf p}^{c_t}=\mathbf E_t\bar{\mathbf p}^{w}$.
Both the single-arm  and bi-manual configuration use horizon $H=30$, at most $M=2$ objects,
$N=30$ points per object, and two grippers. Dimensions and translations
are in meters. The future action index $h\in\{1,\ldots,H\}$ is distinct
from flow-matching time $\tau\in[0,1]$. Depth is not needed in PointPilot.
\subsection{Subtask Planning}
\label{app:planner}

\paragraph{Prompt and task features.}
The VLM receives the current main-view image and the following prompt;
line wrapping is for presentation. No historical image is appended.
\begin{Verbatim}[fontsize=\footnotesize,frame=single,
  framesep=5pt]
Instruction: {instruction}
Previous subtasks:
{previous_subtasks}
Predict the current executable left/right subtask
pair and the immediate next distinct pair that
begins within the next {horizon} actions.
Use NONE for an inactive hand. Return exactly:
<CURRENT>
Left: ...
Right: ...
</CURRENT>
<NEXT>
Left: ...
Right: ...
</NEXT>
If no distinct stage begins in the horizon,
return <NEXT> NONE </NEXT>.
\end{Verbatim}
The four delimiters are atomic special tokens. Only the hidden states
inside \texttt{CURRENT} are passed downstream.
Training uses teacher-forced answers; inference generates an answer and
re-encodes it with the same image and prompt to extract these features.
\texttt{NEXT} contributes language supervision and subsequent memory,
but is not a downstream condition.

\paragraph{Labels and memory.}
Let stage $j$ have interval $[b_j,e_j)$ and canonical left/right label
$s_j$. For effective current stage $k$, CURRENT is $s_k$; NEXT is the
first later distinct label whose start satisfies $t<b_j\le t+H$, or
\texttt{NONE}. Instruction paraphrases are sampled from the first at
most ten stored candidates; stage labels remain canonical.
Training memory is empty for $t<b_0+H$ and otherwise contains
$(s_0,\ldots,s_k)$ after adjacent deduplication, including the current
stage. During rollout, memory instead starts empty and appends predicted
CURRENT followed by NEXT, skipping empty/NONE entries and repetitions
of its last item. Training history is constructed
from GT annotations, not from VLM rollouts; the two history distributions
are not identical.

\subsection{3D Grounding}
\label{app:grounding}

\paragraph{Main-view geometry encoding.}
After Qwen's $2\times2$ merge of $14\times14$ ViT patches, each output
token corresponds to a $28\times28$ region of the resized image. For
source size $(H_0,W_0)$ and resized size $(H',W')$, use
$\widetilde{\mathbf K}=\operatorname{diag}(W'/W_0,H'/H_0,1)\mathbf K$.
On the merged grid $(h_f,w_f)$, location $(i,j)$ uses
$\tilde{\mathbf u}_{ij}=[jW'/w_f,iH'/h_f,1]^\top$, without a
half-cell offset. Its $D=64$ depth hypotheses are
\begin{equation}
\begin{aligned}
 z_d&=z_{\min}+\frac{z_{\max}-z_{\min}}{D(D+1)}d(d+1),\\
 \mathbf x^c_{ij,d}&=z_d\widetilde{\mathbf K}^{-1}\tilde{\mathbf u}_{ij},
 \qquad d=0,\ldots,D-1.
\end{aligned}
 \label{eq:app:rays}
\end{equation}
We use $z_{\min}=0.01$, $z_{\max}=1.60$, and main-frame workspace bounds
$\mathbf a=(-1.10,-0.50,0.00)$, $\mathbf b=(2.25,1.10,1.60)$.
Define $\ell_\epsilon(x)=\log[\max(x,\epsilon)/\max(1-x,\epsilon)]$
after clipping $x$ to $[0,1]$, with $\epsilon=10^{-5}$. The per-token
position encoder is
\begin{equation}
 \mathbf e^{3D}_{ij}=f_{\rm PE}\!\left(
 \operatorname{vec}_{d,xyz}\!\left[
 \ell_\epsilon\!\left(\frac{\mathbf x^c_{ij,d}-\mathbf a}{\mathbf b-\mathbf a}\right)
 \right]\right).
 \label{eq:app:petr}
\end{equation}
Division is elementwise. The shared per-location MLP is implemented as
two $1\times1$ convolutions, $192\rightarrow8192\rightarrow2048$, with
ReLU between them. These are feature-grid encodings, not measured depth.
Out-of-workspace coordinates are clipped, but do not mask valid RGB tokens.

With final-layer VLM image features $\mathbf H^{\rm img}$, CURRENT
features $\mathbf H^{\rm cur}$, and $\phi=\mathrm{Linear}\circ\mathrm{LN}$,
\begin{equation}
\mathbf Z^{\rm img}=\phi(\mathbf H^{\rm img}+\mathbf e^{3D}),\quad
\mathbf Z^{\rm task}=\phi(\mathbf H^{\rm cur}),\quad
\mathbf Z^{\rm ground}=[\mathbf Z^{\rm img};\mathbf Z^{\rm task}].
\label{eq:app:memory}
\end{equation}
Here $\phi$ maps $2048$ to $512$. Both grounding keys and values use
this memory; it contains neither auxiliary-view tokens nor NEXT tokens.

\paragraph{Queries and predictions.}
$Q=2$ learned 3D references $\mathbf r_q$, initialized uniformly in
$[0,1]^3$, provide query positions through a 3D sine/cosine encoding and
an MLP. Query content starts at zero. Six 512-wide decoder layers apply
self-attention with position added to query/key, cross-attention to
$\mathbf Z^{\rm ground}$, and a feed-forward block, each with a residual
connection and post-LayerNorm. Query position is also added to the
cross-attention query. Each layer has eight attention heads and FFN
width 2048. Only the final normalized queries are supervised.

The box head outputs $(\mathbf o_c,\mathbf o_s,\mathbf o_r)$ of widths
$(3,3,6)$, decoded as
\begin{equation}
\begin{aligned}
 \hat{\mathbf c}_q&=\mathbf a+(\mathbf b-\mathbf a)\odot
 \sigma(\mathbf o_{c,q}+\ell_\epsilon(\mathbf r_q)),\\
 \hat{\boldsymbol\eta}_q&=\exp(\mathbf o_{s,q}),\qquad
 \hat{\mathbf R}_q=\operatorname{GS}_{\rm rows}(\mathbf o_{r,q}).
\end{aligned}
 \label{eq:app:box}
\end{equation}
$\operatorname{GS}_{\rm rows}$ normalizes the first 3-vector, orthogonalizes
and normalizes the second, and uses their cross product as the third row.
$\boldsymbol\eta$ denotes full side lengths. The box MLP is
$512\rightarrow512\rightarrow512\rightarrow12$ with ReLU hidden layers.
A separate $512\rightarrow512\rightarrow512\rightarrow4$ head, with
LayerNorm/ReLU hidden layers, predicts probabilities $\pi_{q,r}$ over
invalid, left, right, and shared roles. Inference maximizes total query
log-probability over feasible active-role sets
$\varnothing,\{L\},\{R\},\{S\},\{L,R\}$; unused queries are invalid.
Slot 0 holds a left/shared object and slot 1 a right-exclusive object.

\subsection{Multi-view Geometric Alignment}
\label{app:multiview}

The main and wrist views share one Qwen visual encoder and merger, but
only the main view enters the language transformer. For auxiliary view
$v$, construct rays using its own resized intrinsic matrix as in
Eq.~\eqref{eq:app:rays}. Given current world-to-camera extrinsics
$\mathbf E_{v,t}$ and $\mathbf E_t$, align their metric coordinates by
\begin{equation}
 \bar{\mathbf x}^{c_t}_{v,i,d}
 =\mathbf E_t\mathbf E_{v,t}^{-1}\bar{\mathbf x}^{c_v}_{v,i,d}.
 \label{eq:app:multiview_transform}
\end{equation}
The transform precedes workspace normalization and uses the current
anchor's wrist calibration, not a fixed initial pose. The transformed
coordinates pass through the \emph{same} position encoder in
Eq.~\eqref{eq:app:petr}, with the same main-frame bounds.
For visual-merger output $\mathbf F_{v,i}\in\mathbb R^{2048}$,
\begin{equation}
 \mathbf Z^{\rm aux}_{v,i}=\psi(\mathbf F_{v,i}+\mathbf e^{3D}_{v,i}),
 \qquad \psi=\mathrm{Linear}_{2048\rightarrow512}\circ\mathrm{LN}.
 \label{eq:app:aux_fusion}
\end{equation}
$\psi$ is shared across auxiliary views but separate from $\phi$.
No view-ID embedding is added. Unlike main-view fusion, auxiliary fusion
uses visual-merger features without a language-transformer pass.
These tokens are appended only to DiT conditions, not to grounding memory.
Patch counts follow each view's actual processor grid; padded tokens and
missing views are masked. This geometric branch requires per-view
intrinsics and extrinsics, including moving-wrist calibration.

\subsection{Point Flow DiT}
\label{app:dit}

\paragraph{Anchors and object features.}
For each selected predicted OBB, deterministic Halton coordinates
$\mathbf u_n\in[-\tfrac12,\tfrac12]^3$ use bases $(2,3,5)$ and indices
$n=1,\ldots,N$. Anchors are
\begin{equation}
 \mathbf p_{0,m,n}
 =\hat{\mathbf R}_m
  (\hat{\boldsymbol\eta}_m\odot\mathbf u_n)
  +\hat{\mathbf c}_m.
 \label{eq:app:anchors}
\end{equation}
This is volume sampling, not FPS or depth-surface sampling.
Project the eight OBB corners with the main intrinsic, take their pixel
extrema, and clip the resulting ROI to the image. All corners must lie
in front of the camera; otherwise that downstream object slot is invalid.
The object token is
\begin{equation}
 \mathbf Z_m^{\rm obj}=f_o\!\left[
 f_{\rm PN}(\mathbf P_{0,m});\,
 f_{\rm box}(\mathbf b_m^{2D});\,
 f_{\rm ROI}(\mathbf H^{\rm img},\mathbf b_m^{2D})\right].
 \label{eq:app:object}
\end{equation}
PointNet max-pools a 1024-wide geometry feature; the image-size-normalized
2D box gives 256 features. ROIAlign extracts a $2\times2$ crop from raw
main-view VLM image features \emph{before} PETR addition, then maps it to
512 features. Their 1792-wide concatenation is projected to 512 dimensions.
The grounding-query hidden vector is not added to this token.

\paragraph{Point and gripper tokens.}
For noisy normalized displacement $\mathbf x_{\tau,h,m,n}$, define
\begin{equation}
\begin{aligned}
\mathbf r^p_{h,m,n}
&= W_\Delta \mathbf x_{\tau,h,m,n}
 + b_\Delta
 + f_g(\gamma(\mathbf p_{0,m,n}))
 + \mathbf e^{\mathrm{slot}}_m
 + \mathbf e^{\mathrm{future}}_h, \\[3pt]
\mathbf z^p_{h,m,n}
&= f_p\!\left(
 [\mathbf r^p_{h,m,n};\mathbf e^\tau]
 \right), \qquad
\mathbf z^g_{h,a}= f_g^{\mathrm{in}}\!\left(
 [\mathbf e^{\mathrm{grip}}_a
  + \mathbf e^{\mathrm{future}}_h;
  \mathbf e^\tau]
 \right).
\end{aligned}
\label{eq:app:tokens}
\end{equation}
The 27-wide $\gamma$ concatenates normalized/clipped XYZ and four bands
of sine/cosine features with frequencies $2^j\pi$, $j=0,\ldots,3$.
XYZ normalization uses center $(0.55,0.30,0.72)$ and half extent
$(1.75,0.95,0.90)$, clipped to $[-1.5,1.5]$.
Both token types share a learned future-step table and a sinusoidal
encoding $\mathbf e^\tau$ of $\lfloor1000\tau\rfloor$.
Flow time is \emph{concatenated}, not added or supplied through AdaLN.
The separate input MLPs use LayerNorm, Linear, SiLU, Linear with widths
$1536\rightarrow768\rightarrow768$.
Gripper tokens contain no current gripper state, GT value, or added noise.

Each future step contains $MN$ point tokens followed by two gripper
tokens; 32 learned registers precede the entire sequence. Conditions are
\begin{equation}
 \mathbf C_t=[\mathbf Z^{\rm ground};\mathbf Z^{\rm obj};
                 \{\mathbf Z^{\rm aux}_v\}_v].
 \label{eq:app:condition}
\end{equation}
The 12-block DiT alternates cross-attention and bidirectional
self-attention, with standard pre-layer normalization and a FFN per block.
It has 12 heads of width 64, internal width 768, and FFN width 3072.
After final LayerNorm and projection to 512, separate
LayerNorm--\allowbreak Linear--\allowbreak SiLU--\allowbreak Linear heads output point velocities
$\hat{\mathbf v}\in\mathbb R^3$ and gripper logits $\hat\ell\in\mathbb R$.
Invalid point slots are zeroed and masked as self-attention keys;
their outputs are excluded from the loss. Both gripper tokens remain
valid. Padding uses one batched forward, not individual sample forwards.

\begin{table}[!htbp]
\centering
\setlength{\belowcaptionskip}{5pt}
\caption{Bimanual tensor shapes. $N_v$ is view-dependent ($N_0$ for the main view) and
$L_{\rm cur}$ includes CURRENT-span padding. Sequence length is
$32+30(2\cdot30+2)=1892$. Masks identify valid entries.}
\label{tab:app:shapes}
\small
\setlength{\tabcolsep}{5pt}
\renewcommand{\arraystretch}{1.12}
\begin{tabularx}{\linewidth}{@{}Xl@{}}
\toprule
Tensor & Shape \\
\midrule
Per-view visual/PETR tokens & $[B,N_v,2048]$ \\
Grounding memory & $[B,N_0+L_{\rm cur},512]$ \\
Box outputs / role logits & $[B,2,12]$ / $[B,2,4]$ \\
Anchors / object tokens & $[B,2,30,3]$ / $[B,2,512]$ \\
DiT input & $[B,1892,768]$ \\
Point velocity & $[B,30,2,30,3]$ \\
Gripper logits & $[B,30,2]$ \\
\bottomrule
\end{tabularx}

\end{table}

\paragraph{Inference.}
Starting with Gaussian point noise, use $K=16$ forward-Euler evaluations:
\begin{equation}
 \mathbf x^{(j+1)}=\mathbf x^{(j)}+K^{-1}
 \hat{\mathbf v}_\theta(\mathbf x^{(j)},j/K;\mathbf C_t).
 \label{eq:app:euler}
\end{equation}
For $j=0,\ldots,K-1$, time is bucketed as above. Metric displacement is
$\Delta\hat{\mathbf P}=0.1\mathbf x^{(K)}$.
Gripper logits are recomputed, not integrated; the returned open
probabilities are $\sigma(\hat\ell^{(K-1)})$ from the last evaluation.

\subsection{From Point Flow to Executable Robot Actions}
\label{app:execution}

Convert the anchor and predicted future points to world coordinates
using $\mathbf E_t^{-1}$. For corresponding sets $\mathbf x_n$ and
$\mathbf y_n$, let $\bar{\mathbf x},\bar{\mathbf y}$ be their centroids.
Rigid least-squares fitting has the Kabsch solution
\begin{equation}
\begin{gathered}
\sum_n
(\mathbf x_n-\bar{\mathbf x})
(\mathbf y_n-\bar{\mathbf y})^\top
= U\Sigma V^\top, \\[4pt]
D_R=\operatorname{diag}
\bigl(1,1,\det(VU^\top)\bigr), \\[4pt]
\hat R=VD_RU^\top,
\qquad
\hat{\mathbf d}
=\bar{\mathbf y}-\hat R\bar{\mathbf x}.
\end{gathered}
\label{eq:app:kabsch}
\end{equation}
For the resulting homogeneous transform $\Delta\hat T^w_{h,m}$,
each role-assigned end effector receives
\begin{equation}
 \hat T^w_{a,h}=\Delta\hat T^w_{h,m}T^w_{a,0}.
 \label{eq:app:eef}
\end{equation}
Each future transform is fitted relative to the same anchor, not
accumulated from previous predictions. A shared object applies the
same rigid transform to both hands while preserving their different
anchor offsets. An unassigned hand holds its anchor pose. This mapping
assumes rigid coupling and consumes both translation and rotation.
The RoboTwin adapter rejects fits above 5-cm point RMSE and converts
predicted open probabilities using hysteresis: above 0.7 opens, below
0.3 closes, otherwise retaining the previous command. Poses and gripper
commands are passed to IK/control, followed by a new observation and
planning call after the chunk. No learned EEF-pose head is required.

\subsection{Training Targets and Objectives}
\label{app:objectives}

\paragraph{Targets and stage boundaries.}
Let $T_m^w(t)$ map object-local coordinates to world coordinates.
Training uses Hungarian-matched \emph{predicted} boxes to sample anchors,
then transports them with GT object motion
$\Delta T^w_{h,m}=T_m^w(f_h)[T_m^w(t)]^{-1}$:
\begin{equation}
\begin{aligned}
 \bar{\mathbf p}_{h,m,n}^{c_t}
 &=\mathbf E_t\Delta T^w_{h,m}\mathbf E_t^{-1}
      \bar{\mathbf p}_{0,m,n}^{c_t},\qquad
 \mathbf x_{1,h,m,n}=(\mathbf p_{h,m,n}^{c_t}-\mathbf p_{0,m,n}^{c_t})/0.1.
\end{aligned}
 \label{eq:app:targets}
\end{equation}
For ordinary samples in stage $k$, $f_h=\min(t+h,e_k-1)$, so point
targets hold at the stage endpoint. Gripper labels follow frame $t+h$
inside the stage; beyond it they repeat the next stage's \emph{first}
command, or the terminal command if no next stage exists. Both hands
use $1=$open and $0=$closed. A special boundary anchor at $b_{k+1}-1$
is supervised as stage $k+1$, switching its plan, selected objects, and
future targets together. Short stages are supported by terminal hold.

\paragraph{Planning and grounding.}
Planning loss is mean causal cross entropy over assistant-answer tokens,
including CURRENT/NEXT delimiters; prompt, image, and padding labels
are ignored. For predicted query $q$ and valid GT object $j$, matching uses
\begin{equation}
\begin{gathered}
C_{qj}
= 5\|\hat{\mathbf c}_q-\mathbf c_j\|_1
+ \|\log\hat{\boldsymbol\eta}_q
     -\log\boldsymbol\eta_j\|_1
+ d_R(\hat R_q,R_j)
- \log\pi_{q,r_j}, \\[5pt]
\sigma^*
= \underset{\sigma}{\arg\min}
  \sum_j C_{\sigma(j),j}.
\end{gathered}
\label{eq:app:matching}
\end{equation}
Here, $\sigma$ ranges over injective assignments from
ground-truth objects to predicted queries.
For $A=\hat R R^\top$, the angular error is
\begin{equation}
 d_R=\operatorname{atan2}\!\left(
 \frac{\|\mathbf v(A)\|_2}{2},\,
 \operatorname{clip}\!\left(
 \frac{\operatorname{tr}A-1}{2},-1,1\right)\right),
 \label{eq:app:rotation_error}
\end{equation}
where
$\mathbf v(A)=(A_{32}-A_{23},\allowbreak
A_{13}-A_{31},\allowbreak A_{21}-A_{12})$.
With $J$ matched objects across the batch and unmatched role targets
$\bar r_{bq}=\mathrm{invalid}$, define
\begin{equation}
\begin{aligned}
 L_c&=\frac1{3J}\sum_j\|\hat{\mathbf c}_{\sigma^*(j)}-\mathbf c_j\|_1,\\
 L_s&=\frac1{3J}\sum_j\|\log\hat{\boldsymbol\eta}_{\sigma^*(j)}-
                                    \log\boldsymbol\eta_j\|_1,\\
 L_R&=\frac1J\sum_j d_R(\hat R_{\sigma^*(j)},R_j),\\
 L_{\rm role}&=-\frac1{BQ}\sum_{b,q}\log\pi_{bq,\bar r_{bq}},\\
 L_{\rm ground}&=5L_c+L_s+L_R+L_{\rm role}.
\end{aligned}
 \label{eq:app:ground_loss}
\end{equation}
The assignment is per example; the sums above pool its matched pairs.
Box losses are zero when $J=0$. Center error is measured in meters,
size error in log space, and role loss is unweighted cross entropy.
Unlike the matching cost, center/size losses average over coordinates.

\paragraph{Flow matching and gripper classification.}
For each example, independently sample noise $\epsilon\sim\mathcal N(0,I)$
and $u\sim\mathrm{Beta}(1.5,1)$, and set
$\tau=0.999(1-\min(u,0.999))$. All its tokens share this flow time.
The interpolation and target velocity are
\begin{equation}
 \mathbf x_\tau=(1-\tau)\epsilon+\tau\mathbf x_1,
 \qquad \mathbf v^*=\mathbf x_1-\epsilon.
 \label{eq:app:flow}
\end{equation}
Let $w_i$ mask valid scalar point coordinates and $m^g_j$ mask valid
gripper labels. The action objectives are
\begin{equation}
\begin{aligned}
 L_{\rm pointflow}&=\frac{\sum_i w_i(\hat v_i-v_i^*)^2}
                              {\max(1,\sum_i w_i)},\qquad
 L_{\rm gripper}=\frac{\sum_j m^g_j\operatorname{BCEWithLogits}(\hat\ell_j,g_j)}
                            {\max(1,\sum_j m^g_j)}.
\end{aligned}
 \label{eq:app:action_losses}
\end{equation}
Point validity combines selected-object/projection validity and
future-object validity. Terminal-hold targets remain supervised.
Both grippers are supervised independently of active object count.
Conditions are repeated for two independent noise/time draws per example.
The joint loss is
\begin{equation}
 L=L_{\rm plan}+L_{\rm ground}+L_{\rm pointflow}+L_{\rm gripper}.
 \label{eq:app:total}
\end{equation}
The VLM, including its shared visual encoder, is fully fine-tuned.
Predicted box coordinates are detached before anchor sampling and ROI
projection; action losses therefore update shared feature paths, including
multiview PETR, but not the box head through its geometric outputs.
Matching and discrete role selection are not differentiated. No additional
EEF-pose, rigidity, or intermediate-decoder-layer loss is used.

\section{Experimental Settings}
\label{app:experiments}

\paragraph{Policy interface.}
PointPilot receives the task instruction, current RGB observations, and
generated subtask history, initialized empty at episode start. We use
greedy language decoding, a 30-step prediction horizon, and 16
flow-matching integration steps. Predicted point motion is converted
to end-effector targets by rigid alignment; gripper commands are also
model-predicted. Replanning uses fresh observations after each chunk
of 30 targets, unless execution terminates earlier. 

\begin{table}[t]
\centering
\begin{tabular}{lrr}
\hline
Setting & Trials per task & Execution budget \\
\hline
LIBERO Spatial / Object & 50 & 500 \\
LIBERO Goal / Long & 50 & 1000 \\
RoboTwin clean & 100 & 400 \\
Paired cross-embodiment & 100 per embodiment & 300 \\
Real world & 20 & 600 \\
\hline
\end{tabular}
\caption{Evaluation budgets. Simulation budgets count environment action
calls; physical budgets count predicted waypoint steps, excluding
low-level interpolation. A lower task-specific environment limit takes
precedence. All settings use horizon and replanning chunk length 30.}
\label{tab:app:eval_settings}
\end{table}

\subsection{Simulation Experiments}
\label{app:simulation_settings}
Our model is jointly trained on the Libero and Robotwin 2.0 datasets by 60k steps, using 4*80G A800 GPUs.
\paragraph{LIBERO.}
We evaluate Spatial, Object, Goal, and LIBERO-10 (Long) using a single
main-view RGB observation and a wrist view observation. Fixed episode initializations are shared
across policies and identified by task, demonstration, and initial
simulator state in the evaluation manifest. Ten zero-action settling
steps precede the first policy observation and do not consume the
budget in Table~\ref{tab:app:eval_settings}. Success uses the environment's
task predicate. We use MuJoCo 3.3.2 and robosuite 1.4.0 with the
corresponding LIBERO assets.

\paragraph{RoboTwin.}
We evaluate the 50 tasks (100 trials per task) under the clean configuration with
the ALOHA-AgileX embodiment and head, left-wrist, and right-wrist RGB
views. Candidate seeds are examined in ascending order from 100000;
acceptance requires both
successful expert planning and satisfaction of the task success
predicate. The environment is then reset before evaluating the learned
policy. We use SAPIEN 3.0.0b1 and MPLib 0.2.1 with clean task assets;
randomized-background results are not mixed into this protocol.

\subsection{Real-World Experiments}
\label{app:realworld_settings}

\paragraph{Platform and tasks.}
The physical platform is a CobotMagic bimanual system with an external
camera and two wrist cameras. The three RGB streams are aligned to a
common observation timestamp. We consider QR-code scanning and sorting,
cube-to-basket transfer, and pouring glass beads between cups. Task-specific
policies use demonstrations with stage annotations, camera-frame object
tracks, and binary gripper commands. Each task is trained by 20k steps, using 50 demonstrations.

\paragraph{Execution and assessment.}
Object arrangements are fixed before policy comparison and independently
reset for each trial. A robot-side client converts model predictions
using camera-to-base calibration and robot kinematics, with feedback
checks before execution. Success requires the complete task: scanning
all specified QR codes and placing the scanned objects in the designated
tray, placing and releasing the cube in the basket, or transferring the
beads into the receiving cup without spilling. Operator corrections
are not permitted; safety interventions and policy-induced control
failures count as unsuccessful trials. Offline point-flow visualizations
are not counted as physical task successes.

\subsection{Zero-Shot Cross-Embodiment Transfer}
\label{app:cross_embodiment_settings}

\paragraph{Frozen policy and paired scenes.}
We transfer an ALOHA-trained policy to ARX-X5 and dual-Piper without
target-embodiment demonstrations for task fine-tuning
at evaluation time. The same checkpoint, prompt, point-flow sampling,
and learned gripper prediction are used for source and target robots.
Calibration, robot descriptions, kinematics, and execution interfaces
are embodiment-specific. Each source--target comparison uses 100
accepted scenes and equal execution budgets.

Scene pairing preserves object assets and arrangements, table/background
configuration, main-camera calibration, and the exact instruction.
Equality is checked after natural settling
and any robot initialization adjustment, using tolerances of 2\,mm in
object position and $2^\circ$ in orientation. Robot appearance and wrist
observations remain embodiment-dependent. For Piper, initial end-effector
poses are additionally aligned to the ALOHA reference within 2\,mm and
$2^\circ$ per arm. ARX retains its native feasible initialization;
identical initial end-effector poses are not assumed for that comparison.

\paragraph{Feasibility and evaluation.}
Task eligibility requires the target embodiment's task assets and a
feasible expert execution. Scenes are accepted only when
both source and target experts succeed and the pairing checks pass;
Piper also requires successful initial-pose alignment. Experts qualify
the initial conditions but do not provide actions or intermediate
labels to the policy. The accepted list is frozen before policy testing
and shared by the ALOHA reference and corresponding target policies.
The metric therefore measures transfer on expert-feasible paired scenes,
not arbitrary target initializations.
\end{document}